\documentclass[graybox]{svmult}

\usepackage{mathptmx}       %
\usepackage{helvet}         %
\usepackage{courier}        %
\usepackage{type1cm}        %
\usepackage{makeidx}         %
\usepackage{graphicx}        %
\usepackage{multicol}        %
\usepackage[bottom]{footmisc}%

\usepackage[utf8]{inputenc}
\usepackage[T1]{fontenc}

\usepackage{graphicx,float,import,hyperref,placeins,microtype}%
\usepackage{color,multirow,stackengine}
\usepackage{listings}
\usepackage{amsmath,amssymb} %

\newcommand{\talker}[1]{{\tt\small [#1]}}
\newcommand*\samethanks[1][\value{footnote}]{\footnotemark[#1]} 
\usepackage[colorinlistoftodos,prependcaption,textsize=tiny]{todonotes}
\usepackage{colortbl}

\makeindex             %

\begin{document}

\title*{A Talker Ensemble: the University of Wrocław's Entry to the NIPS 2017 Conversational Intelligence Challenge}
\titlerunning{A Talker Ensemble: UWr's entry to NIPS 2017 Conversational Int. Challenge}
\author{Jan Chorowski\thanks{Author order is alphabetical and does not reflect authors' contributions.}, Adrian Łańcucki\samethanks, Szymon Malik\samethanks, Maciej Pawlikowski\samethanks,\\Paweł Rychlikowski\samethanks, Paweł Zykowski\samethanks} 
\authorrunning{Chorowski, Łańcucki, Malik, Pawlikowski, Rychlikowski, Zykowski}
\institute{Institute of Computer Science\\
                   University of Wrocław \\
                   Joliot-Curie 15, 50-383 Wrocław, Poland \\}
\maketitle

\abstract*{
We present Poetwannabe, a chatbot submitted by the University of Wrocław
to the NIPS 2017 Conversational Intelligence Challenge,
in which it ranked first ex-aequo.
It is able to conduct a~conversation with a~user in a~natural language.
The primary functionality of our dialogue system is context-aware
question answering (QA),
while its secondary function is maintaining user engagement.
The chatbot is composed of a number of sub-modules,
which independently prepare replies to user’s prompts
and assess their own confidence.
To answer questions, our dialogue system relies heavily
on factual data, sourced mostly from Wikipedia and DBpedia,
data of real user interactions in public forums,
as well as data concerning general literature.
Where applicable, modules are trained on large datasets using GPUs.
However, to comply with the competition's requirements,
the final system is compact and runs on commodity hardware.}

\abstract{
We present Poetwannabe, a chatbot submitted by the University of Wrocław
to the NIPS 2017 Conversational Intelligence Challenge,
in which it ranked first ex-aequo.
It is able to conduct a~conversation with a~user in a~natural language.
The primary functionality of our dialogue system is context-aware
question answering (QA),
while its secondary function is maintaining user engagement.
The chatbot is composed of a number of sub-modules,
which independently prepare replies to user’s prompts
and assess their own confidence.
To answer questions, our dialogue system relies heavily
on factual data, sourced mostly from Wikipedia and DBpedia,
data of real user interactions in public forums,
as well as data concerning general literature.
Where applicable, modules are trained on large datasets using GPUs.
However, to comply with the competition's requirements,
the final system is compact and runs on commodity hardware.}

\section{Introduction}
The NIPS 2017 Conversational Intelligence Challenge promoted creation of conversational agents (chatbots), that could maintain a conversation with a human peer about a given news or encyclopedic article. Each conversation was opened with an article excerpt presented to both parties, after which they could communicate freely by asynchronously exchanging text messages.
In this contribution we present the University of Wrocław's chatbot called Poetwannabe.

The objective of the competition was to achieve high conversation quality which was evaluated by human judges. Such formulation gave chatbot designers much freedom in the system design.
To simplify the problem, we have set a
goal for our chatbot as enhancing the user's
understanding of terms and concepts relevant to the discussed article.
Therefore, the bot was primarily built to answer factual questions, propose follow-up topics,
and serendipitously present interesting facts related to the article.
What distinguishes Poetwannabe from a search engine is maintaining a constant
dialogue interaction with the user.
Thus, our secondary goal was responding to common utterances, regardless of the background article.
To address this need, our system implements a general conversational module,
supported by a large corpus of publicly available dialogues
from on-line forums.

In general, we assumed asymmetry between the bot and the user: the bot is useful to its users and focuses on answering their needs.
This assumption allows us to fully utilize the strength of computer programs,
such as their super-human ability to index and memorize large bodies of data.
It follows from this assumption, that the chatbot should not examine user's understanding on the contents of the article.
Our interpretation of the rules means that, by design, the bot is easy to tell apart from a human, and would be a poor contestant in the Turing test.
\section{Background and Related Work}

Conversational systems spur the imagination since the inception
of Alan Turing’s famous paper on the ``Imitation Game'',
now known as the Turing test \cite{turing}.
To pass the test, the agent has to 
deceive a human judge into thinking they talk to another person. Throughout the years, the test evolved into various competitions, the oldest and one of the most notable of them being the Loebner Prize\footnote{\url{http://www.aisb.org.uk/events/loebner-prize}}, introduced in 1991. Thus far, no bot managed to win the one-time-only prize, awarded to a program able to truly fool the judges.

Early successful dialogue systems follow a \textit{pattern-response} paradigm. We can point to ELIZA~\cite{eliza} as a prime example of this approach.
The idea for ELIZA was to simulate a psychoanalyst, who reflects
interlocutor’s statements in their responses.
ELIZA incorporates linguistic rules, keyword matching and ranking
as well as “memory”. Another famous chatbot, A.L.I.C.E.~\cite{wallace2009anatomy}, or Alicebot, can be thought of as an updated
and more developed version of ELIZA.
In Alicebot those rules are written in an XML dialect
called Artificial Intelligence Markup Language (AIML). 

Participants of the recent conversational intelligence competitions,
including the NIPS 2017 Conversational Intelligence
Challenge and the Alexa Prize\footnote{\url{https://developer.amazon.com/alexaprize}},
have established the \textit{ensemble} paradigm, in which independent responses
collected from different modules are combined by
a response selection policy.
Poetwannabe follows this broad paradigm, and shares many traits with
other submissions to the above competitions.
Two examples of similar chatbots are MILABOT \cite{montreal_bot},
an Alexa Prize contestant, and Bot\#1337 \cite{1337bot} 
the second winner of NIPS 2017 Conversational Challenge.
MILABOT, created by a team from Montreal Institute for Learning Algorithms
is an ensemble of multiple agents, including deep learning ones,
as well as variants of ELIZA and Alicebot.
It chooses its responses using a dialogue manager trained with
reinforcement learning on rewards received from users. Bot\#1337,
created by the Moscow Institute of Physics and Technology,
implements a set of skills (chit-chat, QA, topic detection,
text summarization, etc.) and employs a dialogue manager
trained in a supervised setting to select appropriate responses.

Poetwannabe utilizes a similar ensemble-based architecture,
although instead of training a dialogue manager,
we have employed confidence scores calibrated on selected dialogues.

\section{System Description}

The design of our chatbot is modular.
Several subsystems,
called \emph{talkers}, independently monitor the conversation state
and propose responses.
This modularity made for easy development and parallelization of talkers, with minimum coordination
between talker developers. Moreover, it enhanced the reliability of our chatbot:
failures or timeouts of individual talkers can degrade particular responses,
but the chatbot is guaranteed to always reply
within the allocated time budget. The modular approach also
introduces some downsides: it is difficult to endow the bot
with a consistent persona, and to fine-tune the policy,
which selects the best response among
those proposed by the talkers.

The talkers can be broadly grouped into three categories by their
responsibility: question answering,
fact generation and general conversation handling.
The QA talkers achieve the main objective of question answering.
The fact generation talker displays to the user facts
that are relevant, but missing from the initial article.
The general conversation talkers respond to prompts
in a way that creates the illusion of having a real conversation
with an intelligent peer.
They also respond to users' quirks, that we have observed
during the preliminary stage of the competition.
For instance, curious users investigate if the chatbot
speaks in languages other
than English, or if it can evaluate mathematical expressions.
Finally, we include an open-source implementation
of the Alicebot
as a fall-back mechanism.
We list all talkers in Table~\ref{talkerstable}
and their exemplary response proposals in Table~\ref{talkersexample}. 

\setlength{\tabcolsep}{5pt}
\begin{table}[t]
	\centering
    \caption{Talkers grouped by role}
    \label{talkerstable}
    \begin{tabular}{| p{0.28\linewidth} | p{0.65\linewidth} |}
        \hline
        \rowcolor[gray]{.9}
        \multicolumn{2}{| l |}{\textbf{Question answering}}\\
        \hline
        Wikipedia QA & Paragraph indexing, neural QA trained on large dataset\\
        \hline
        DBpedia fact indexer & Fuzzy triple matching using word embeddings\\
        \hline
        Simple Wikipedia definitions & Uses extracted definitions\\
        \hline
        Article topic finder & Article summarization and noun phrase extraction, finding similar Wikipedia titles\\
        \hline
        \rowcolor[gray]{.9}
        \multicolumn{2}{| l |}{\textbf{Fact generation}}\\
        \hline
        Simple Wikipedia facts & Extracts interesting sentences and scores their similarity to the given article\\
        \hline
        \rowcolor[gray]{.9}
        \multicolumn{2}{| l |}{\textbf{General conversation handling}}\\
        \hline
        Matching utterance embeddings & NN search over single utterances and pairs (Wikiquote, Reddit comments, handcrafted)\\
        \hline
        Trivia questions & Matched using word embeddings\\
        \hline
        Alice chatbot & Avoid boring answer by rewarding new terms\\
        \hline
        Misc behaviors & Handling non-english prompts and math\\
        \hline
    \end{tabular}    
\end{table}

\setlength{\tabcolsep}{3pt}
\begin{table}[t]	
    \caption{Example of talkers' responses}
    \label{talkersexample}
    
    \textbf{Article}: Cyclone Monica degraded

	\noindent\textit{Cyclone Monica was expected to hit Darwin as a category five storm, with winds of up to 350 km/h (220 mph). But the national weather bureau downgraded it to category two on the morning of April 25, 2006, when it lost power after making landfall. Monica was a category five cyclone when it touched down in the remote Aboriginal community of Maningrida, in the Northern Territory, late on 24th night.}

	\noindent\textit{A spokeswoman for the Northern Territory Police, Fire and Emergency Services said some parts of the area had been damaged by the storm. Maningrida "certainly suffered extensive damage to some buildings and structures," she told the Sydney Morning Herald, although no serious injuries have been reported.}\\
    
	\noindent \textbf{User}: What continent did cyclone Monica impact? \\

    \centering
    \begin{tabular}{l l p{0.69\linewidth}}
        Wikipedia QA & 1.12 & I'd say Australia.\\
        DBpedia &  0.46 & A continent is one of several very large landmasses on Earth.\\
        Wikiquote & 0.41 & A life is not important except in the impact it 	has on other lives.\\
        Alice & 0.40 & That's not something I get asked all the time.\\
        Simple Wikipedia & 0.23 & A continent is a large area of the land on Earth that is joined together.\\
        Simple facts & 0.22 & Interesting fact: Monica is a female given name.\\
        Trivia & 0.20 & I know the answer, do you: What theme is central to the movies The Lost Weekend, The Morning After and My Name Is Bill W.??\\
        Forum comments & 0.08 & Like this guy surrounded by ignorance overcame it all.\\
        Topic guess & 0.01 & It's about the remote Aboriginal community of Maningrida.\\
        Popular dialogues & 0.01 & What is it that you want to know?\\
        Abacus & 0.00 & I have no idea.\\
        Gimmick & 0.00 & Well...\\
    \end{tabular}
\end{table}

\subsection{Detailed Response Generation Algorithm}\label{sec:algo}
\begin{figure}[t]
  \centering
    \includegraphics[width=0.4\textwidth]{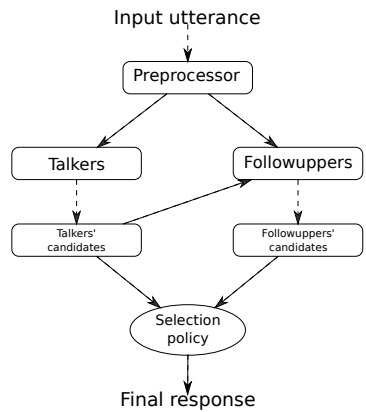}
  \caption{\small{Response generation algorithm. The user utterance first goes through a common preprocessing pipeline, which involves spell checking and coreference resolution. Then, the talkers can propose their responses. Afterward, the talkers' replies are modified during the follow-up stage. Finally, the set of candidate responses and follow-ups is ordered by the confidence and a response is selected to be shown to the user.}}
  \label{fig:response_gen}
\end{figure}

In addition to being able to choose the best reply from the candidate list calculated by independent subsystems, we wanted to enable aggregation of responses
coming from several talkers. For example, it makes sense to combine the answer to the user's query
from the Wikipedia QA Talker with a related Simple Wikipedia fact. 
Those considerations led us to a two-round response selection algorithm
(Figure~\ref{fig:response_gen}).
During the proposal round, all talkers independently
propose candidate responses and state their confidence scores.
Next, during the follow-up round,
each talker can inspect the ranked list of candidate responses, and freely modify its reply.
The new proposal can, for example, be an extension of the old one.
Again, all talkers execute their follow-up 
methods independently and update their confidence scores.
Versatility of the follow-up mechanism
allowed us to easily aggregate responses
of multiple talkers (e.g., responding to a user question and adding
a related factoid statement), as well as to implement complex response selection rules
for certain user prompts. 

We note that since the follow-up mechanism
can combine responses of several talkers, we do not keep track
of which talker was active in which turn of the conversation.
Instead, the talkers are assumed to record
the state of the conversation themselves, and respond whenever
the past dialogue matches their triggering patterns.
If a talker wants to be active during several consecutive dialogue turns,
it can indicate it to the dialogue system by responding with a high confidence.
For instance, QA talkers propose answers with high confidence if they classify user's utterance as a question, and with low confidence otherwise. Conversly, general dialogue talkers typically reply with a medium confidence.

\subsubsection{Common Utterance Pre- and Post-processing}
Each user utterance undergoes a common set of transformations:
we first spell-check it with Hunspell\footnote{\url{https://hunspell.github.io/}},
making sure that we do not replace words that are in the vocabulary lists
of the QA modules.
We then tokenize it and resolve coreferences using CoreNLP~\cite{corenlp}.
This form is passed to the talkers. Finally, each chatbot response
is capitalized and filtered for foul language. We have used an extensive
vocabulary blacklist, although we have decided to whitelist words
entered by the user.
This way the chatbot is able to respond to sensitive topics,
but cannot initiate them by itself.

\subsection{Detailed Talker Descriptions}

\subsubsection{Simple Wikipedia Talker (SWT)}

Analysis of the data from the preliminary competition round
revealed that users, when given a background article,
tend to ask about definitions.
We identify such queries with a heuristic function,
which considers presence of common asking phrases
(\emph{tell me about, what is}, etc.) accompanied by pronouns,
as well as the location of single-word prepositions.
Two talkers independently recognize query questions and assess their confidence
score with this heuristic: Simple Wikipedia Talker and DBpedia Talker. This section describes the former.

Simple Wikipedia Talker responds with either: the opening sentences
of a Simple Wikipedia (SW) article, or a definition from Wiktionary.
It assigns the highest confidence score when the question contains SW article title,
or a Wiktionary term.
Otherwise, it uses a built-in off-line search engine to search for
the definitions of related entities, and retrieves the opening sentence
of the best matching SW article.
When SWT is highly confident about its response,
it also suggest other related topics.
If the user happens to ask about them later on,
SWT gives a big boost to its confidence score
to maintain a coherent conversation.
Example suggestions are shown in Table~\ref{swtfollowups}.

\setlength{\tabcolsep}{5pt}
\begin{table}[ht]
    \centering
    \caption{Examples of SWT suggestions}
    \label{swtfollowups}
    \begin{tabular}{| p{0.28\linewidth} | p{0.65\linewidth} |}
        \hline
        \rowcolor[gray]{.9}
        \textbf{Wikipedia title} & \textbf{Suggested topics} \\\hline
        hamburger & kfc, barbecue, hot dog, french fries, maple syrup \\\hline
        mathematics & number theory, logic, set theory, number, geometry \\\hline
        england & wales, bristol, yorkshire, uk, sheffield \\\hline
        mozart & sonata, counterpoint, accompaniment (music), symphony, virtuoso \\\hline
        marie curie & pierre curie, irène joliot-curie, dmitri mendeleev, ivan pavlov, george gamow \\\hline
        software & personal computer, computer game, computer program, computer hardware, apple macintosh \\\hline
        european union & euro, social democracy, european commission, european union budget, euroscepticism \\\hline
        sandstone & limestone, shore, sediment, silt, sand \\\hline
        hamlet & othello, twelfth night, english literature, the importance of being earnest, falstaff \\\hline
        the lord of the rings & the hobbit, middle-earth characters, luke skywalker, jedi, middle-earth \\\hline
    \end{tabular}    
\end{table}

\paragraph{\textbf{Calculating SW Match Quality}}
Given an article, we find the best follow-ups using cosine similarity between
embeddings of Wikipedia article titles. Those embeddings are pre-computed
using Word2vec~\cite{mikolov2013distributed}.
For a Wikipedia page, we try to predict what entries it links to,
and what entries link to it.
We used an artificially created corpora of tuples \texttt{(title, linked\_title)}.
Every tuple consisted of an article title, and a title of another article it links to.
Word2vec algorithm was used to embed links and titles, like if they were
ordinary words.

\paragraph{\textbf{Wikipedia Indexer}}\label{indexer}
Our Wikipedia indexer can either return a list
of paragraphs rated by their similarity to the query,
or an entire Wikipedia article containing the most relevant paragraph.
The first option is usually used when dealing with a user's question.
The second one comes in handy when 
we just want to know more about the subject, in which case we feed the entire main article into the search engine.

We gather all unigrams and bigrams from the query. Wikipedia paragraphs
are then rated based on the number of occurrences
of those phrases in them. We use inverse document frequency (idf)
scores of $n$-grams to accentuate the presence of the most important words.
Idf of a phrase is the percentage of documents in the corpus,
that phrase occurs in. The scores were computed on Wikipedia texts.
Due to RAM limitations, indexes are stored in cached databases on the hard drive.

\subsubsection{Wikipedia QA}
Our primary question answering module combines our implementation
of FastQA~\cite{weissenborn2017fastqa} with a Wikipedia indexer described above.
A similar idea is proposed in~\cite{Chen2017ReadingWT}.
The~indexer finds a~set of Wikipedia paragraphs most related to a user's question,
and scores their relevancy to the~topic. FastQA then proceeds to find an~answer
in each of them. The~best response is chosen using the~quality
of a given paragraph according to the~indexer and the~confidence
returned by FastQA. The module usually processes
between 20 and 40 passages per question,
within the competition time limit.

Answers to many questions can also be found in the~article that started
the conversation. To make use of that, we form a~set of on-topic excerpts
at the~beginning of the conversation. It contains the~article as well as
fragments of a~Wikipedia page which most resembles the~article.
Those paragraphs help answer questions regarding the~main subject.
However, we do not use them if the~question is not similar enough with
the~article.

Wikipedia QA rephrases the question before performing a~Wikipedia search.
The first verb occurring after a~question word (\emph{what, when, who,} etc.)
is placed just after the~first noun phrase, which follows that verb.
The form of a~verb is changed to be grammatically correct.
For example, \emph{when did the~war start} is transformed into
\emph{when the~war started}.
This heuristic slightly improved the~quality of retrieved paragraphs.

The talker usually has problems answering simple definition questions
like \emph{Who was Niels Bohr?}
We adjust confidence scores of our talkers so that SWT and DBpedia talker
would take over on this kind of queries.
If the~highest idf word in the query still has low idf,
we lower the confidence score of Wikipedia QA to refrain it
from participating in chit-chat conversations.

We trained the model on SQuAD training set \cite{Rajpurkar2016SQuAD10},
using 300d GloVe word embeddings~\cite{pennington2014glove}. Our training process was similar to that of \cite{weissenborn2017fastqa}. 
Best model was achieved after 9 epochs. 
We used early stopping to stop the training after epoch 21.
Unfortunately we were not able to replicate the~result from \cite{weissenborn2017fastqa}. Our best model got only $72.34$ F1 on SQuAD development set.
For comparison, \cite{weissenborn2017fastqa} report $77.07$ F1,
which was slightly below $80.6$ F1 presented by the top model from that time \cite{rnet}.

\paragraph{\textbf{Negative Answers Experiment}}
In SQuAD, each question comes with a background document that contains the correct answer.
Thus, a FastQA model trained on SQuAD often responds nonsensically when presented an off-topic text.
We have performed an experiment to check if FastQA can be trained to
not only point out the correct span in a fitting article (answer positively),
but also decide if the~span is missing from the text (answer negatively).
Although it did not yield fully satisfying results, we briefly describe our approach.

We have prepared additional datasets for training to answer negatively. Questions from SQuAD training set have been coupled with paragraphs from Wikipedia found by the indexer. Most of them did not contain the correct answers. Analogous pairs were created for the~SQuAD validation set, and used to test the models. A~special \texttt{neg} token has been added at the~end of every paragraph. The model could point to it, in order to answer negatively. The embedding of \texttt{neg} was learned. A~perfect model would discard all off-topic contexts, while still finding correct answers in paragraphs that contain them. By training on additional examples from Wikipedia, we were able to reach 87\% accuracy in discarding negative paragraphs. Unfortunately, this came with a severe drop in performance on the primary task: the model could only achieve 58 F1 on SQuAD validation set. Ultimately we found the~trade-off too costly, and our final system does not use this feature.

\subsubsection{DBpedia Talker}
This talker answers questions, based on the knowledge in the form of
RDF triples extracted from English Wikipedia infoboxes.
Due to memory limits, we only utilize the top 100\,000 resources
according to the unofficial
DBpedia PageRank\footnote{\url{http://people.aifb.kit.edu/ath/\#DBpedia_PageRank}}.

Each triple consists of a resource identifier, an attribute identifier, and a value of that resource's attribute. The same attributes are common for many resources, and one resource attribute can have many values. At the time of writing, the newest dataset contains information extracted from Wikipedia in October 2016.

DBpedia Talker tries to identify the topic of user's question among the known resources. In case of success, it looks for connections between parts of a question and facts about the topic. The best matching facts are formed into declarative sentences, e.g., \emph{Albert Einstein birth date is 1879-03-14}, or \emph{Christopher Columbus was an Italian explorer, navigator, colonizer, and citizen of the Republic of Genoa.}

Resource names are stored in a trie. Given a tokenized question, we try to match subsequences of tokens with known identifiers.  Only one resource is returned. We prefer the ones with longer prefix match or higher PageRank. To optimize this step, we omit fragments which do not look like resource identifier (\emph{who is}, \emph{how old}, etc.), or would produce too many results, like \emph{the}.

The crucial step is to find relations between the question and attributes. For this task we use word embeddings together with the English language thesaurus. In addition to checking words appearing in the query, we also look at their neighborhoods (in a sense of cosine distance) and synonyms. For example, when answering a question \emph{Who was Albert Einstein's wife?} it turns out the word \emph{wife} is closely related to the word \emph{spouse}, and \emph{spouse} is an attribute name of a formal relationship with a woman.

The connections are scored and the best one is chosen. We answer by combining the resource name, attribute name, and values of the attribute. The response to the question about Einstein's wife would be \emph{Albert Einstein spouses are Mileva Marić, Elsa Löwenthal}. We have a dedicated submodule for handling age questions (\emph{How old...}, \emph{How many years...}, etc.). The talker returns the value of an \emph{abstract} attribute when no interesting connections were found.

DBpedia Talker assigns partial confidence scores throughout the entire process. The final value is a combination of scores obtained during resource finding, connections finding and response building. Resource finding confidence comes from the matching quality and the PageRank of a given resource. The other partial scores depend mostly upon the numbers of connections retrieved and sentences generated, as well as the Levenshtein distance between synonyms and matched attributes.

\subsubsection{Topic Guess Talker}

During the~first stage of the~competition we noticed that users
often expected Poetwannabe to provide a~short summary of the~article.
This talker is supposed to fill that role. Using phrases picked
from the competition preliminary stage dataset,
we handcrafted a~small database of questions regarding general topic
of the~text. An~example of such query would be
\textit{What is the~article about?}
User prompt is softly matched with those phrases using word vectors,
and if the~match is good enough, Topic Guess Talker answers with a high confidence.

We utilize a~simple frequency-based extractive text summarizer
to give a~one-sentence overview of the~article.
We also gather the~most interesting noun phrases from the~passage,
and use them to provide short answers like \emph{The~text is about <noun phrase>}.
In addition, we try to find a~Wikipedia title most closely related to the~topic.
If the~indexer manages to get a~close match, this title forms another short reply.
From all these proposition we choose one and return it as a~final response.

\subsubsection{K-NN Talker}
Large datasets of on-line dialogue records are scarce,
especially those regarding a particular topic.
In contrast, there is an abundance of publicly available conversations
on forums concerning various entities like news,
articles, blog posts, videos, pictures, etc.
We focus on forum comment sections.

The character of comments differs from that of a conversation between two
individuals. With many participants, comments typically form a tree in
which an utterance is a node, and an edge signifies a response. Those trees
tend to be rather shallow and wide. Even though a given path in the tree
might not be a coherent dialogue, pairs and triplets
of consecutive turns can form a short but logical chat.

We propose a talker that exploits local coherence of comment sections and treats
each pair of subsequent comments $(A, B)$ as a single training sample. All pairs from
the corpus are embedded as pairs of 300d vectors $(E_A, E_B)$,
and stored in a database.
For user utterance $U$, we retrieve
a pair $(A, B)$ with the highest cosine similarity $\cos(E_A, E_U)$,
and respond with $B$. The data for this module comes from a publicly available
dump of Reddit
forums\footnote{\url{https://bigquery.cloud.google.com/dataset/fh-bigquery:reddit_comments}}
and a \texttt{chatterbot-corpus}
Python package\footnote{\url{https://github.com/gunthercox/chatterbot-corpus}}.
We also experimented
with the Hacker News corpus\footnote{\url{https://cloud.google.com/bigquery/public-data/hacker-news}},
but found the responses to be too technical to be used in  a chatbot.
We use two independent K-NN Talkers for Reddit and \texttt{chatterbot-corpus}.

Let $S=(w_1,w_2,\ldots,w_n)$ be a sentence of length $n$, and $E_{w_i}$
a pre-trained embedding of word $w_i$~\cite{mikolov2013distributed}.
We embed sentence $S$ as a Bag-of-Words Embedding (BoWE)
\[E_S = \frac{\sum_i \log{tf(w_i)}E_{w_i}}{\lVert\sum_i \log{tf(w_i)}E_{w_i}\rVert_2},\]
where $tf(w)$ denotes term frequency of word $w$ in the corpus.
In addition, when computing a BoWE
we drop out-of-vocabulary words
and greedily couple together neighboring words to form long phrases
(e.g., consecutive words \emph{Michael} and \emph{Jordan} are embedded as \emph{Michael\_Jordan}).

\paragraph{\textbf{Censorship}}
Reliance on comments scraped from Internet forums poses the obvious risk
of unconsciously using foul language, hate speech, or
touching on controversial and delicate topics.
Filtering out such toxic utterances is a challenging task itself,
and requires some level of comprehension of the utterance and its context.

We quickly found that using a list of forbidden words, which are mainly curse words,
is not sufficient. We trained a binary discriminator neural network,
tasked with rating the ``toxicity'' of an utterance.

A pair $(Q,A)$ was converted to a 600d vectorial
representation, %
where both utterances were embedded as 300d bags-of-words.
Due to lack of data, we trained the toxicity predicting network in a semi-supervised fashion.
We used a small corpus of curse words and banned terms and phrases.
If at least one utterance in a $(Q,A)$ pair contained a forbidden entity,
the pair was labeled as positive (i.e., toxic) example.
A version of such utterance pair with prohibited terms removed
was added to the training set as yet another positive example.
Our conjecture is that sentences with toxic words are,
with high probability, present in toxic contexts.

To facilitate the censorship task, we trade precision over recall and select
utterance pairs, which are rated as being toxic with probability $<0.4$.

\subsubsection{Wikiquote Talker}

The idea behind using Wikiquote was to create a module that would provide
general responses with rare words,
that could be blindly put into the conversation and still fit in.
Quotes meet this condition and fit into a lot of scenarios,
giving Poetwannabe a witty character.
For instance, when the user says \textit{I feel lucky},
Wikiquote talker could respond with a quote
\textit{Luck, that's when preparation and opportunity meet}
(by~Pierre Elliott Trudeau).
We heuristically chose about 50\,000 quotes from Wikiquote to form a database
of possible responses.

The talker embeds each utterance as a tuple
of dense and sparse vectors $(v_{\textit{d}},v_{\textit{s}})$.
Each dense vector $v_{\textit{d}}\in\mathbb{R}^{300}$
is a bag of GloVe embeddings of words from the utterance.
A sparse vector $v_{\textit{s}}\in\mathbb{R}^{|V|}$ 
has a field for each word $w$ in vocabulary~$V$,
set to $\textit{tf}\times\textit{idf}$ for words present in the utterance, and $0$ for absent.
Similarity between pairs of dense and sparse vectors
$(u_{\textit{d}},u_{\textit{s}})$ and
$(v_{\textit{d}},v_{\textit{s}})$,
is calculated as a linear combination of their cosine similarities
$\alpha\cos(u_{\textit{d}},v_{\textit{d}}) + (1-\alpha)\cos(u_{\textit{s}},v_{\textit{s}})$.
The talker has an internal state, which stores dense and sparse
representations of both the context of a dialogue and the article.

Context vectors are initialized to zero and updated each turn,
using the current user sentence and the last bot response.
We focus on the latest utterances,
but we do not completely forget earlier parts of the conversation.
The article is summarized as a bag of interesting words,
which are chosen heuristically.
Context vectors are combined with article vectors to produce dense
and sparse summary of the dialogue.

We keep track of rare words occurring in both the dialogue and the article.
When matching the best response, we prioritize those words, and
the influence of the article naturally decays as the conversation progresses.
To diversify the answers we randomly sample the reply among top scoring candidates.
To avoid random topic changes, we lower the influence of utterances containing too many unknown words.
The final response is also penalized if it is too similar to the user query,
or if the query has only common words.

\subsubsection{Alice Talker}

This talker fulfills two objectives: maintains the~conversation flow
with reasonable generic responses to user's questions,
and sustains a coherent personality of the chatbot.
At the~beginning of the~conversation, we choose values for bot predicates
describing its personality
(gender, name, birthplace, age, hobbies, favorite color, and many others)
-- these values will be occasionally used in bot responses.
Alice talker uses off-the-shelf AIML rules from A.L.I.C.E.~\cite{wallace2009anatomy}.
From the~perspective of our code, AIML chatbot is treated as a black box talker.
Generally speaking, every chatbot can be used in this way, provided it has
a confidence assessment mechanism.

Since AIML rules do not provide a clear notion of the rule's fitness
to a~given utterance, Alice talker scores its confidence
with the following principles:
\begin{itemize}
  \item Shorter Alice responses are generally more natural (and preferable).
  \item Alice talker should not repeat itself during the~conversation.
  \item Accurate responses are preferred (e.g., \emph{It is Paris, of course}
        is better than generic reply like \emph{Please, go on},
        since the~former is clearly related to the user's question).
\end{itemize}
  
This ``originality score'' is computed in the~following way:
we created a~corpora of Alice conversations in which the~user utterances
were taken randomly from Wikiquote or from NIPS Conversation Challenge data.
For every word generated by Alice, we computed its counts. The~final score
depended on the~logarithmic~count of the least common term introduced by the chatbot
(i.e., contained in chatbot's answer, and not present in the user's utterance).
Note that we had to use only terms introduced by a bot.
Some general rules repeat fragments of user's prompts,
which can contain very specific words.

\subsubsection{Simple Fact Generator}

Given the~article, this talker tries to present some facts related to it.
All facts are self contained Simple Wikipedia (SW) sentences.
To obtain them, we perform a~number of SW searches,
and gather interesting sentences among the results.

We try to find definitions of phrases occurring in the~article.
In addition, we search for a~SW entry most closely resembling the~article.
Then, we use a~simple heuristic: interesting sentences on a~SW page contain part
of the~title of that page, and do not have any pronouns occurring before it. Result of this filtering can be observed in the example below.\\

\begin{lstlisting}[breaklines]
GOOD: Alan Turing was born in Maida Vale, London. 
GOOD: Turing went to St. Michael's, a school at 20 Charles Road, St Leonards-on-sea, when he was six years old.
BAD: His father was part of a family of merchants from Scotland. 
BAD: His mother, Ethel Sara, was the daughter of an engineer.
\end{lstlisting}

During the~conversation, Simple Fact Generator presents these facts,
starting from the~most interesting ones.
When rating a~response, we consider the~following questions:
\begin{itemize}
  \item Do any key words suggest the~user wants to learn about the~main topic?
  \item Does a~fact come from a~highly scored article? Is it the~first sentence of it?
  \item Does it contain words used previously by the~user?
  \item Does it have proper length?
\end{itemize}

In order to prevent this talker from dominating the~discussion,
the~confidence score is penalized if Simple Fact Generator already spoke.

\subsubsection{Trivia Questions}
The Trivia Questions talker was designed to ask potentially interesting questions
to the user. Instead of building a general question asking model we have opted
to use higher quality trivia question lists that can be found on the Internet.
This way the chatbot would not leave the impression of drilling the user
with a reading comprehension test.
Instead, it has a possibility to entertain the user. 

We have gathered a set of trivia questions found on Internet portals:
Moxquizz\footnote{\url{http://moxquizz.de/download.html}},
Tat's Trivia\footnote{\url{http://tatarize.nfshost.com/}}
and an archive of the Irc-wiki\footnote{\url{https://web.archive.org/web/20150323142257/http://irc-wiki.org}}.
The talker matches questions to the dialogue by computing
a tf-idf weighting of terms in all candidates.
Next, it  expresses each Q-A pair as a single real-valued vector by computing
a tf-idf weighted average of GloVe vectors for all words in that pair.
It then computes similar vectors for the past user utterances and uses
the cosine distance to select the best matching question.
Finally, the talker samples its confidence score based on the quality
of the best matching answer, to make sure that the questions will not be triggered
each time a certain set of keywords was entered by the user.

\subsubsection{Miscellaneous Talkers (Abacus, Gimmick)}
These talkers were inspired by the~analysis of dialogues from preliminary phase
of the~competition, as well as tests conducted during public events at our faculty.
In order to add a~nice touch to our system and correctly handle
some non-standard situations, we designed a~couple of small specialized rule-based mechanisms.
They can detect and evaluate mathematical expressions,
respond to greetings in different languages, and react to urls,
e-mail addresses and emoji characters sent by the~user. 

Mathematical expressions can be embedded within an utterance,
and possibly be written partly in English, e.g., \emph{What is 2 plus 2}.
These are parsed and evaluated heuristically with regular expressions.
For urls and email addresses we randomly sample one of several handcrafted responses.

In order to handle non-English greetings we took advantage of multilingual sentences from Tatoeba\footnote{\url{tatoeba.org}} and trained a~language classification model\footnote{\url{https://github.com/saffsd/langid.py}}. An utterance is classified as a~foreign salutation if it is not in English, and has common words with any greeting from Tatoeba. If this rule is satisfied for more than one language, we check whether the~language returned by the classifier is among them.

\subsection{Balancing and Prioritizing Talker Confidences}
Poetwannabe selects the final response based on talkers' self-assessments of confidence.
More complicated selection rules and special cases were facilitated
by the mechanism of follow-ups (described in section \ref{sec:algo}).
Each talker computed its confidence score based on linguistic features
of the user's prompt, the context of the conversation, and its internal state.
To balance the confidences and to ensure that the talkers respond to utterances
they were designed to handle, we have prepared a dialogue corpora containing users'
prompts of several types (chit-chat, questions about definitions, general questions,
offensive dialogues). We have then linearly scaled the confidence scores
such that the talkers most relevant to each corpus are also the ones
with the highest confidence scores on it.
Additionally, the follow-up mechanism was used to jointly promote
QA talkers for user prompts that were identified as questions.

\section{Conclusions and Future Work}

We are pleased with the outcome of our work,
although the problem is far from solved.
We see limitations of this system and possible improvements.
Chatbots in general, including our system, struggle with maintaining rich context
of the conversation. We suspect a solution to this problem would cause
a noticeable boost in performance.

Dividing the chatbot into independent talkers proved convenient from the architectural
point of view, but this decision has clear downsides.
Selecting the best response (i.e., adjusting the confidence scores)
turned out to be challenging. There definitely is a great potential
in developing a more robust supervisor of talkers.
Independence of individual subsystems resulted in our bot having a multitude of different,
sometimes even contradictory, personalities. Advanced interactions with the user,
like asking them to provide some details, also were more difficult to accomplish. 

Our question answering module would benefit greatly from a large QA dataset
where the context is not strictly necessary to find an answer.
Queries from SQuAD are often very context-dependent.
Questions like \emph{Who was the medical report written for?}
or \emph{How many did this epidemic in China kill?} simply cannot be answered
without an access to a specific paragraph. If not for this characteristic,
we might be able to train a better Wikipedia indexer, which would increase
the likelihood of finding the right passage.

It is worth mentioning that during the last few months the top performing models
on SQuAD dataset almost matched
human performance\footnote{\url{https://rajpurkar.github.io/SQuAD-explorer/}}.
Having access to a better QA architecture would also bring an improvement
to the system. On top of that, presenting the content of the article
to the user could be made easier with more intelligent summarizer.

We found that data-driven modules, such as K-NN Talker,
can carry a more engaging casual conversation than classic,
rule-based mechanisms like Alicebot. However, in our opinion,
conversational bots work best when combining both options.

Using large amounts of raw text from the Internet turned out to be quite problematic.
Sanitizing the data and ensuring our bot would not say anything inappropriate
was a small challenge in and of itself. Similarly, making the system
run on low resources with all the knowledge we wanted it to process,
called for compromises.

Lastly, chatbots and conversational user interfaces still need to find their optimal use cases. The Conversational Intelligence Challenge stimulated answering this question by defining the necessary boundary conditions: a dialogue format and relevant quality measures. Beyond these, contestants were granted free choice in defining the main purpose of their chatbots. We decided on a rather narrow purpose for our chatbot, defined as the user's utility. Thus our bot was designed to be help the user understand the article by providing an elaborate contextual question answering mechanism and only basic general conversational abilities. Thus the bot was tasked not to impersonate a human, but to provide a conversational interface to bot's knowledge. However, there exist other viable chatbot goals, such as trying to act like a human and pass the Turing test.  We believe that the jury is still out on the question of the purpose of conversational agents.

\begin{acknowledgement}
The authors want to thank the NIPS 2017 Conversational Intelligence Challenge for organizing the competition, and especially Valentin Malykh for help with resolving technical issues with chatbot deployment. We thank our labmates Adam Kaczmarek and Michał Zapotoczny for not complaining about our loud brainstorming sessions. %
We also thank the University of Wrocław for providing a travel grant. Work on neural generators and QA systems was partially supported by Poland National Science Center under grant NCN Sonata 276459 2014/15/D/ST6/04402. Finally, we would like to thank everyone who was patient enough to talk to our chatbot during its development, for providing useful feedback and ideas.
\end{acknowledgement}
\section*{Appendix: Sample Dialogues}
\addcontentsline{toc}{section}{Appendix}
We present a set of curated dialogues that best showcase all talkers.
\\

\newcommand{\afterarticleskip}{.5em}
\newcommand{\betweendialoguesskip}{2em}
\newenvironment{hanginglines}[1][1em]
  {%
   \renewcommand\\{\par\hangindent#1}%
   \parindent0pt %
   \hangindent#1 %
  }
  {\par}

\noindent{\bf Article}: During the Second World War, Turing worked for the Government Code and Cypher School (GC\&CS) at Bletchley Park, Britain's codebreaking centre that produced Ultra intelligence. For a time he led Hut 8, the section which was responsible for German naval cryptanalysis. Here he devised a number of techniques for speeding the breaking of German ciphers, including improvements to the pre-war Polish bombe method, an electromechanical machine that could find settings for the Enigma machine. Turing played a pivotal role in cracking intercepted coded messages that enabled the Allies to defeat the Nazis in many crucial engagements, including the Battle of the Atlantic, and in so doing helped win the war.
\vspace{\afterarticleskip}

\begin{hanginglines}
\noindent{\bf User}: hello, let us chat! \\
{\bf Bot}: Hi there!. I already started. \talker{Alice} \\
{\bf User}: what role did Alan Turing play in cracking german coded messages? \\
{\bf Bot}: Wait a second... that enabled the Allies to defeat the Nazis in many crucial engagements? \talker{Wikipedia~QA}\\
{\bf User}: why did Allies defeat Nazis? \\
{\bf Bot}: Based on what I know to attack their lines of communications. \talker{Wikipedia~QA} \\
{\bf User}: who won the Battle of the Atlantic? \\
{\bf Bot}: I would say Turing. \talker{Wikipedia QA} \\
{\bf User}: Maybe in a metaphorical sense... \\
{\bf Bot}: Sometimes it is an illustration of what is happening. \talker{K-NN} \\
{\bf User}: when was the Battle of the Atlantic? \\
{\bf Bot}: Maybe the answer is 1866. Well... A cypher (or cipher) is an algorithm for performing encryption (coding) or decryption (decoding). \talker{Wikipedia QA + Simple Fact Generator}\\
{\bf User}: what happened in 1866 \\
{\bf Bot}: It has to be the police caught the murderers. \talker{Wikipedia QA} \\
{\bf User}: is it a historical event, are you sure? \\
{\bf Bot}: There is only one thing in your life YOU can be sure of. \talker{Wikiquote} \\
{\bf User}: which one? death or taxes? \\
{\bf Bot}: Nothing is more important in the face of a war than cutting taxes.\talker{Wikiquote} \\
{\bf User}: Let us come back to the article, ok? \\
{\bf Bot}: Well, this is an interesting subject: the Second World War.\talker{Topic Guess} \\
{\bf User}: how many submarines were in German Navy during the Second World War. \\
{\bf Bot}: It has to be 8. \talker{Wikipedia QA} \\
{\bf User}: Not more? Are you sure? \\
{\bf Bot}: Indeed I am. \talker{K-NN}]
\end{hanginglines} 
\vspace{\betweendialoguesskip}

\noindent{\bf Article}: "I Like Chopin" is a song by Italian singer Gazebo. It was produced in Italy in 1983 by Pierluigi Giombini. It is generally regarded as one of the most significant Italo disco tracks of the mid-1980s. It peaked at number one in several European national charts. The music was composed by Giombini (contrary to popular belief without themes from Chopin), the lyrics were written by Gazebo.
\vspace{\afterarticleskip}

\begin{hanginglines}
\noindent{\bf User}: buongiorno! \\
{\bf Bot}: Ciao! Parlate inglese? It's so much I know in Italian :(  Let's speak in English from now on, shall we? \talker{Miscellaneous} \\
{\bf User}: who was the author of the song "I like Chopin"? \\
{\bf Bot}: An author is narrowly defined as the originator of any written work and can thus also be described as a writer. \talker{DBPedia} \\
{\bf User}: who composed the song "I like Chopin"? \\
{\bf Bot}: I guess Pierluigi Giombini. \talker{Wikipedia QA} \\
{\bf User}: who was Chopin? \\
{\bf Bot}: Fryderyk Franciszek Chopin (1 March 1810 – 17 October 1849) was a Polish-French composer and virtuoso pianist. \talker{Simple Wikipedia} \\
{\bf User}: what Chopin composition do you like best? \\
{\bf Bot}: I don't have a favorite Chopin composition. My favorite band is Pink Floyd. \talker{Alice} \\
{\bf User}: can you play piano?  \\
{\bf Bot}: I know what you mean. It's always nice to just -get- the music. \talker{K-NN} \\
{\bf User}: i can play guitar \\
{\bf Bot}: Do you mostly play Electric or Acoustic? \talker{K-NN} \\
{\bf User}: acoustic \\
{\bf Bot}: There are many interesting things related to it. Say "piano rock", or "gibson sg". \talker{Simple Wikipedia} \\
{\bf User}: what is gibson sg? \\
{\bf Bot}: Gibson SG is a solid body electric guitar model that was introduced in 1961 (as Gibson Les Paul) by Gibson Guitar Corporation. Maybe we can talk about "slide guitar" or "rock (music)"?  \talker{Simple Wikipedia} \\
{\bf User}: No thanks. But may I have the last question? \\
{\bf Bot}:  You have my thanks...\talker{Wikiquote} \\
{\bf User}: who composed Eine Kleine Nachtmusik? \\
{\bf Bot}: I believe it's Mozart. \talker{Wikipedia QA} 
\end{hanginglines}
\vspace{\betweendialoguesskip}

\noindent{\bf Article}: A child prodigy is a child who is very smart at an early age. They can do things that average people can only do when they are adults. Child prodigies are often musicians or mathematicians. A child prodigy does not always grow up to be a world-famous genius. Some of them lose their exceptional gifts in adulthood.
\vspace{\afterarticleskip}

\begin{hanginglines}
\noindent{\bf User}: what is a child prodigy? \\
{\bf Bot}: In psychology research literature, the term child prodigy is defined as a person under the age of ten who produces meaningful output in some domain to the level of an adult expert performer. \talker{DBpedia} \\
{\bf User}: what can do child prodigies? \\
{\bf Bot}: I am quite sure it's calculating skills. \talker{Wikipedia QA} \\
{\bf User}: could you name some musicians who were child prodigies? \\
{\bf Bot}: Biologically, a child is a human being between the stages of birth and puberty. \talker{DBpedia} \\
{\bf User}: I know. But are there any child prodigies playing music? \\
{\bf Bot}: How, specifically, do you know? No I don't think there are any. \talker{Alice} \\
{\bf User}: You are wrong. Consider Mozart for example. \\
{\bf Bot}: No, I am right. Can you think of another example? \talker{Alice} \\
{\bf User}: Franz Liszt for example. \\
{\bf Bot}: Can you think of another example? World is strange... A child prodigy is a child who is very smart at an early age. \talker{Alice + Simple Fact Generator}
\end{hanginglines}
\vspace{\betweendialoguesskip}

\noindent{\bf Article}: Python is an interpreted high-level programming language for general-purpose programming. Created by Guido van Rossum and first released in 1991, Python has a design philosophy that emphasizes code readability, and a syntax that allows programmers to express concepts in fewer lines of code, notably using significant whitespace. It provides constructs that enable clear programming on both small and large scales.
\vspace{\afterarticleskip}

\begin{hanginglines}
\noindent{\bf User}: what is this article about?  \\
{\bf Bot}: The topic of this text is Guido van Rossum.\talker{Topic Guess} \\
{\bf User}: what else is this article about? \\
{\bf Bot}: Created by Guido van Rossum and first released in 1991, Python has a design philosophy that emphasizes code readability, and a syntax that allows programmers to express concepts in fewer lines of code, notably using significant whitespace. \talker{Topic Guess} \\
{\bf User}: It is rather boring. Can we talk about something else? \\
{\bf Bot}: This text is about high-level programming language for general-purpose programming. \talker{Topic Guess}  \\
{\bf User}: I said: something else! \\
{\bf Bot}: It depends. \talker{K-NN}\\
{\bf User}: Depends on what? \\
{\bf Bot}: Good point, I should put what I like in the description. \talker{K-NN}\\
{\bf User}: Which programming language uses significant whitespace? \\
{\bf Bot}: I am convinced that it's Python. Ever seen that one: R. Kelly sings: 'If I can see it then I can do it, if I just believe it, there's nothing to it'. What's the song title?? \talker{Wikipedia QA + Trivia Questions}
\end{hanginglines}

\bibliographystyle{spmpsci}
\bibliography{refs}

\end{document}